\documentclass[sigconf]{acmart}

\usepackage{amsmath,amsfonts}
\usepackage{algorithm}
\usepackage{array}
\usepackage{textcomp}
\usepackage{stfloats}
\usepackage{url}
\usepackage{verbatim}
\usepackage{graphicx}

\usepackage{multirow}
\usepackage{booktabs}
\usepackage[noend]{algpseudocode}  
\usepackage{subfloat}

\AtBeginDocument{%
  \providecommand\BibTeX{{%
    \normalfont B\kern-0.5em{\scshape i\kern-0.25em b}\kern-0.8em\TeX}}}

\setcopyright{acmlicensed}
\copyrightyear{2018}
\acmYear{2018}
\acmDOI{XXXXXXX.XXXXXXX}

\acmConference[Conference acronym 'XX]{Make sure to enter the correct
  conference title from your rights confirmation emai}{June 03--05,
  2018}{Woodstock, NY}
\acmISBN{978-1-4503-XXXX-X/18/06}

\copyrightyear{2024} 
\acmYear{2024} 
\setcopyright{acmlicensed}\acmConference[WWW '24 Companion]{Companion
Proceedings of the ACM Web Conference 2024}{May 13--17, 2024}{Singapore,
Singapore}
\acmBooktitle{Companion Proceedings of the ACM Web Conference 2024 (WWW
'24 Companion), May 13--17, 2024, Singapore, Singapore}
\acmDOI{10.1145/3589335.3651540}
\acmISBN{979-8-4007-0172-6/24/05}

\begin{document}

\title{Object-level Copy-Move Forgery Image Detection based on Inconsistency Mining}


    \author{Jingyu Wang}

\authornote{Both authors contributed equally to this research.}
    \authornote{This work is done when Jingyu Wang is a visiting student at Digital Trust Centre at Nanyang Technological University.}

\author{Niantai Jing}
\authornotemark[1]
\affiliation{%
  \institution{Ocean University of China}
  \city{Qingdao}
  \country{China}
}

\author{Ziyao Liu}
\affiliation{%
 \institution{Nanyang Technological University}
 \country{Singapore}}

\author{Jie Nie}
\authornote{Corresponding author}
\affiliation{%
  \institution{Ocean University of China}
  \city{Qingdao}
  \country{China}}
\email{niejie@ouc.edu.cn}

\author{Yuxin Qi}
\affiliation{%
  \institution{Shanghai Jiao Tong University}
  \country{China}
  }

\author{Chi-Hung Chi}
\affiliation{%
  \institution{Nanyang Technological University}
  \country{Singapore}}

\author{Kwok-Yan Lam}
\affiliation{%
  \institution{Nanyang Technological University}
  \country{Singapore}
  }

\renewcommand{\shortauthors}{Jingyu Wang et al.}
\begin{abstract}
  In copy-move tampering operations, perpetrators often employ techniques, such as blurring, to conceal tampering traces, posing significant challenges to the detection of object-level targets with intact structures.                Focus on these challenges, this paper proposes an Object-level Copy-Move Forgery Image Detection based on Inconsistency Mining (IMNet). To obtain complete object-level targets, we customize prototypes for both the source and tampered regions and dynamically update them. Additionally, we extract inconsistent regions between coarse similar regions obtained through self-correlation calculations and regions composed of prototypes. The detected inconsistent regions are used as supplements to coarse similar regions to refine pixel-level detection.      We operate experiments on three public datasets which validate the effectiveness and the robustness of the proposed IMNet.
\end{abstract}





 \maketitle

\section{Introduction}
The development of network technology makes the image, the important carrier of information, can be spread effortlessly.    Criminals maliciously alter the content of images and disseminate fake news will threaten cyber security.     Copy-move tampering is a common operation in the process of forgery, defined as the process of copying an area in an image (the source region) and pasting it into another area in the same image (the tampered region).                       Since the tampered object has the same illumination and contrast as the other regions in this image, it is more challenging to copy-move forgery image detection than other forgery operations.                       In addition, to make the tampered image more real and better express the misleading content, the forger usually carries out some post-processing operations on the tampered image, such as blurring, rotation, etc., which undoubtedly brings difficulties to the detection of the tampered image.

Copy moving forgery image detection is to locate the source and tampered region in the image which essence is to mine the high similarity area. 
In recent years, copy-move forgery detection based on deep learning has been widely studied \cite{verma2023survey}.      Focus on obtaining a fine-grained detection mask with complete structure, Zhu et al. proposed ARNet \cite{2020AR}, which first obtains a coarse similarity mask and then refines the mask through a process of feature extraction and recovery.      Liu et al. \cite{2022Liu} used the SuperGlue to refine the mask by using the key point matching method after the approximate locating.      The above only detects similar regions through pixel-level modelling leading to the structural integrity of the detected objects not being guaranteed.

To obtain the complete object-level detection target as well as a fine-grained boundary between the object and the background, we propose IMNet in the copy-move forgery detection task.                    IMNet first obtains the feature representation of coarse similar regions through the self-correlation operation.                    Secondly, we customize the source region prototype (SRP) and tampered region prototype (TRP) representations respectively.                    The SRP is updated by inconsistency mining between the TRP and coarse similar region features.             Also, the TRP is updated according to the SRP and coarse similar region features.                    Through this implicit adaptive iterative updating, the SRP and the TRP prototypes can approach the final object-level feature representation.                    Then, we use the Inconsistency Mining module to extract the difference between the coarse similar region and the updated prototype, which is set as the supplementation to the coarse similar region to obtain refined detection results.     The contributions of this paper are as follows:
 
\begin{itemize}
    \item  We propose a copy-move forgery detection framework assisted by the SRP and TRP to detect similar targets with object-level integrity.
\item We propose the Inconsistent Mining module to extract differences in feature representations which are used both in updating the prototype and obtaining the supplementary to refine the coarse similar regions.

\item We conduct comparison as well as ablation experiments on three public datasets to verify the superiority of our proposed IMNet and the effectiveness of each module.
\end{itemize}

\floatname{algorithm}{Algorithm} 
\renewcommand{\algorithmicrequire}{\textbf{Input:}} 
\renewcommand{\algorithmicensure}{\textbf{Output:}} 

 \begin{figure*}
\centerline{\includegraphics[width=16cm]{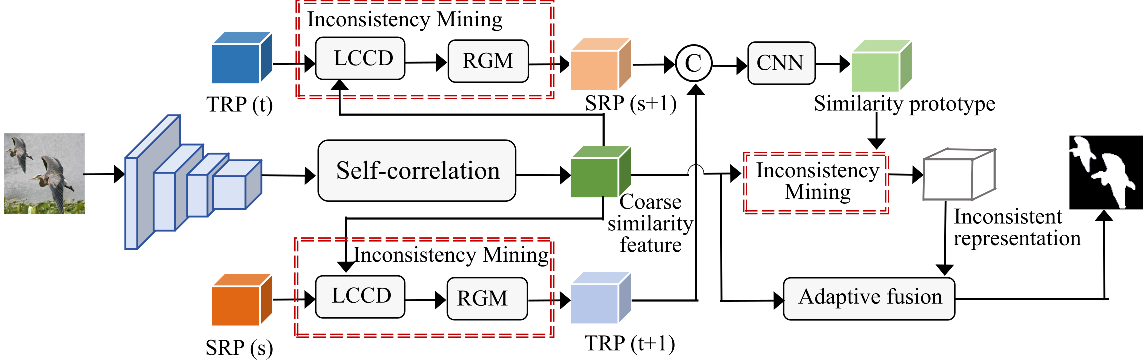}}
\caption{The input of the IMNet is the copy-move forgery image.   The suspect region is obtained to refine the detection effect by mining the inconsistency of a pair of custom prototypes and the coarse similar regions.}
\label{fig:structure}
\end{figure*}

\section{Related work}

In recent years, the deep learning-based copy-move forgery detection method \cite{wu2018image, 2020Dense, 2021CMSDNet, 2020Copy, WANG2024103685} has been concentrated on by researchers due to its superior generalization and ability to capture intricate content compared with the traditional methods \cite{Pun}. Wu et al. \cite{2018Buster} proposed BusterNet, which locates two similar regions through self-correlation calculation and percentage pooling. Based on BusterNet, Zhu et al. \cite{2020AR} proposed AR-Net, performing further encoding and decoding operations to refine detection after obtaining coarse similarity features. Liu et al. \cite{2022Liu} initially locate the target region through object detection, then optimize pixel-level targets through key-point detection. Islam et al. \cite{2020DOA} proposed DOA-GAN using a second-order attention mechanism, which first locates the fuzzy target region and then obtains the final detection result by modeling the global correlation of the initial target. Zhang et al. \cite{CNNT} proposed an integration method of CNN and transformer modules to exchange local and global information to refine the detection. Additionally, Zhao et al. \cite{zhao2024camu} proposed CAMU-Net, highlighting significantly similar regions through coordinate attention. Although the above methods achieve refinement of the detection target to some extent, they still face challenges in accurately distinguishing the target and background boundaries to ensure the integrity of the detected object.  Therefore, we propose IMNet to realize object-level modelling through custom the SRP and TRP and obtain sceptical positions for adjusting coarse similar regions.
\section{Method}
\subsection{Overall structure}
We propose IMNet to ensure target integrity through object-level modelling while obtaining the refined result.    Firstly, the copy-move forgery image is input into the IMNet, and the coarse similarity features are obtained through the CNN feature extraction layer and the self-correlation operation.

We customize TRP and SRP, respectively. The TRP (t) and the SRP (s) are leveraged for inconsistency mining with the coarse similarity feature to obtain the updated SRP (s+1) and TRP (t+1). It is worth noting that in the TRP, the tampered region is significant while the other regions are suppressed. In the coarse similarity feature, both the source and tampered region are significant while the background is suppressed. Therefore, the SRP (s+1) can be obtained by mining the local inconsistency between them. We use a similar method to obtain the TRP (t+1) and fuse it with the SRP (s+1) to generate a pair of similar prototypes.

To further refine the targets, we optimize the coarse similar region feature with a pair of prototypes as auxiliary information.            We use inconsistency mining to obtain a representation of the inconsistency between the two and identify doubtful pixels that are difficult to divide into targets or backgrounds.    The refined detection results are obtained through the adaptive fusion of these doubtful pixels and coarse similarity features.            The overall process of IMNet is shown in Figure \ref{fig:structure}.

\subsection{Inconsistency Mining module}
We propose an Inconsistency Mining module consisting of two modules: Local Correspondence Correlation Detection (LCCD) and Reverse Gate Mechanism (RGM) operation.    This operation has two functions in this paper: (1) Get updated SRP and TRP to optimize object-level object representation;   (2) Detect suspicious areas which are used to refine coarse similarity feature.

LCCD is used to detect the correlation of pixels in the corresponding spatial position.   We calculate from two perspectives: treating a space position as a whole and considering different channels.  The specific calculation method is shown in Algorithm 1.

\renewcommand{\thealgorithm}{1} 
    \begin{algorithm}
        \caption{Local Correspondence Correlation Detection} 
        \begin{algorithmic}[1] 
            \Require{Feature ${F_{x}} \in {\mathbb{R}^{H\times W\times C}}$, ${F_{y}} \in {\mathbb{R}^{H\times W\times C}}$}
            \Ensure{Local correspondence correlation feature ${F_{z}} \in {R^{H\times W\times C}}$}
            \State $V\{\}$, $M\{\}$, $W\{\}$
            \State ${F_{x}} \in {\mathbb{R}^{HW\times C}} \leftarrow Reshape(F_x)$ 
            \State ${F_{y}} \in {\mathbb{R}^{HW\times C}} \leftarrow Reshape(F_y)$
            \For{$k=1:C$}    
                 \For{$l=1:HW$}
                       \State ${F_{k}} = {F^{(l,k)}_{x}} {F^{(l,k)}_{y}}$
                       \State $V\{\}=V\{\}$ insert $F_k$ in length dimension
                \EndFor
                \State $M\{\}=M\{\}$ insert $V\{\}$ in channel dimension
            \EndFor
        \For{$l=1:HW$}
                       \State ${F_{s}} = {F^{(l,.)}_{x}} {F^{(l,.)}_{y}}$
                       \State $W\{\}=W\{\}$ insert $F_s$ in length dimension
 \EndFor

        \State $M\{\}=M\{\}$ insert $W\{\}$ in channel dimension

            \State ${F_{z}} \in {\mathbb{R}^{H\times W\times C}} \leftarrow Reshape (M\{\}\in {\mathbb{R}^{HW\times (C+1)}})$
        \end{algorithmic}
    \end{algorithm}

After obtaining the correlation that represents the local corresponding spatial position, we obtain complementary information through the RGM module, that is, the inconsistent representation.   The operation of RGM is shown in equ (\ref{equ}).
\begin{equation}
\label{equ}
 F^{'}_{z}  = F_z \otimes (1-Gate(F_z))
\end{equation}
where $\otimes$ denotes the multiplication of corresponding elements, and $Gate()$ is implemented by the Sigmoid function.

 \begin{figure*}
\centerline{\includegraphics[width=15cm]{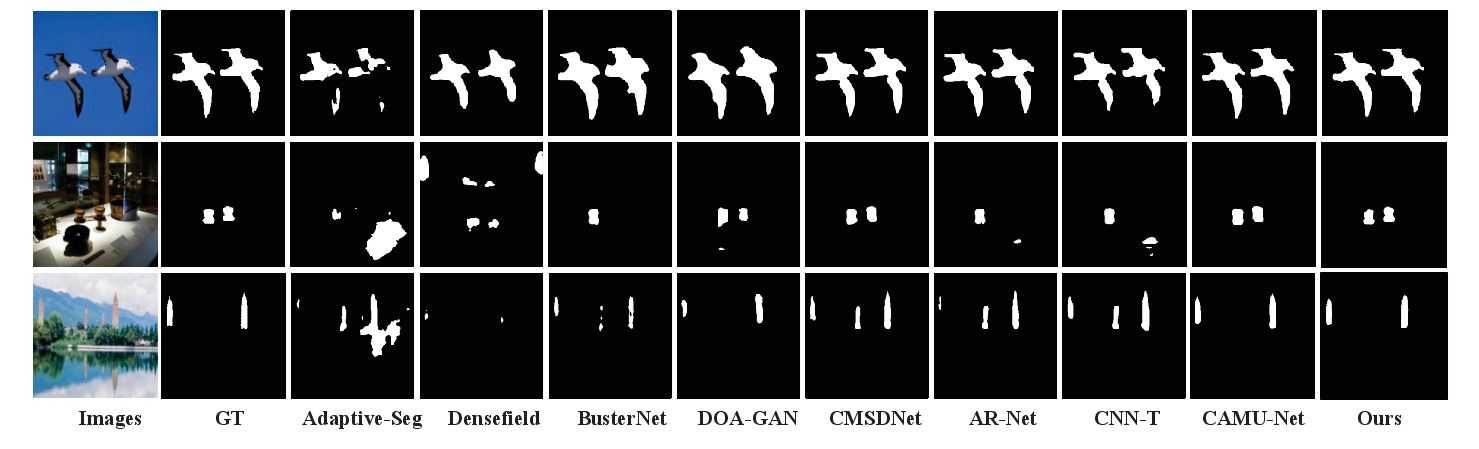}}
\caption{Comparison of detection performance of different algorithms. The first column is to copy-move forgery images, the second column is groundtruth, and the third to ninth columns are mask generated different methods}
\label{fig:view}
\end{figure*}

\section{Experiments}
\subsection{Training data and details}
We train, validate, and test the model on the USC-ISI CMFD dataset \citep{2018Buster} at a ratio of 8:1:1. In addition, we conduct experimental tests on CASIA v2.0 \citep{CASIA}, and the CoMoFoD \citep{CoMoFoD} datasets to prove the effectiveness and strong robustness of our proposed IMNet. All experiments are conducted using the PyTorch framework on NVIDIA 16GB Tesla V100 Gpus. We evaluate the IMNet by precision, recall and F1 metrics.

\begin{table}[!b]
		\renewcommand{\arraystretch}{1.0}
		\caption{The comparison experiments result on the USC-ISI CMFD dataset in precision, recall, and F1 score metrics.}
		\label{tab:uscisi}
		\centering
		\setlength{\tabcolsep}{3mm}{
			\begin{tabular}{ c c c c}
				\toprule
				\multirow{1}*{Methods}&Precision  &Recall&F1-score \\
				
				\midrule
				\multirow{1}*{BusterNet \citep{2018Buster}}& 54.84 & 43.50 & 44.59  \\
				\multirow{1}*{DOA-GAN \citep{2020DOA}} & 80.95 & 72.01 & 74.15 \\
				
                \multirow{1}*{AR-Net \citep{2020AR}} & 73.53 & 75.61 & 72.42 \\
                \multirow{1}*{CNN-T \citep{CNNT}} & 80.25 & 79.18 & 79.82 \\
                \multirow{1}*{CAMU-Net \citep{zhao2024camu}} & 55.80 	 & 41.60 	&47.60 \\
                
				\midrule
				\multirow{1}*{Baseline}                  & 72.68	&74.45	&75.58 \\
				\multirow{1}*{Prototype}                 & 75.95	&76.06	& 76.69 \\
				\multirow{1}*{prototype (update)}        & 76.97	&77.84	&77.87 \\
				\multirow{1}*{Spatial}                    & 79.11	&79.08	&79.30 \\
				\multirow{1}*{Channel}                & 80.39	&80.54	&80.67 \\
				\multirow{1}*{IMNet}                     & \textbf{82.05}	&\textbf{82.13}	 & \textbf{82.16} \\
				\bottomrule 
		\end{tabular}}
	\end{table}

\subsection{Contrast experiment}

    We select the state-of-the-art methods for copy-move forgery detection and compare them with our proposed IMNet in terms of experimental data and visualization effects, which fully verifies the effectiveness of IMNet.    Comparison methods include Adaptive-Seg \citep{Pun}, Densefield \citep{Cozzolino}, BusterNet \citep{2018Buster}, DOA-GAN \citep{2020DOA}, CMSDNet\citep{2021CMSDNet}, AR-Net \cite{2020AR}, CNN-T \citep{CNNT}, and CAMU-Net \citep{zhao2024camu}.

\begin{table}[!b]
		\renewcommand{\arraystretch}{1.0}
		\caption{The comparison experiments result on the CASIA CMFD dataset in precision, recall, and F1 score metrics.}
		\label{tab:casia}
		\centering
		\setlength{\tabcolsep}{3mm}{
			\begin{tabular}{ c c c c}
				\toprule
				\multirow{1}*{Methods}&Precision  &Recall&F1-score \\
				
				\midrule
               
				\multirow{1}*{Adaptive-Seg \citep{Pun}} & 23.17 & 5.14 & 7.42\\
				\multirow{1}*{DenseField \citep{Cozzolino}} & 24.92 & 26.81 & 25.43\\
				\multirow{1}*{BusterNet \citep{2018Buster}} & 42.15 & 30.54 & 33.72\\
				\multirow{1}*{DOA-GAN \citep{2020DOA}} & 54.70 & 39.67 & 41.44\\
                \multirow{1}*{AR-Net \citep{2020AR}}       & 54.36 & 50.90 & 48.03\\
                \multirow{1}*{CNN-T \citep{CNNT}}          & 50.48 & 51.05 & 50.76\\
				\multirow{1}*{CAMU-Net \citep{zhao2024camu}} & \textbf{55.80} 	& 	41.60 	 	&47.60\\
                
				\midrule
				\multirow{1}*{Baseline}                  & 46.82	&43.98	&45.08 \\
				\multirow{1}*{Prototype}                 & 48.71	&45.33	&46.69 \\
				\multirow{1}*{Prototype (update)}        & 50.87	&46.84	&47.92 \\
				\multirow{1}*{Spatial}                    & 53.18	&48.41	&49.57 \\
				\multirow{1}*{Channel}                & 54.65	&49.09	&50.38 \\
				\multirow{1}*{IMNet}                     & 55.47	&\textbf{51.81}	 & \textbf{52.47} \\
				\bottomrule 
		\end{tabular}}
	\end{table}

\begin{table}[!b]
		\renewcommand{\arraystretch}{1.0}
		\caption{The comparison experiments result on the CoMoFoD dataset in precision, recall, and F1 score metrics.}
		\label{tab:comofod}
		\centering
		\setlength{\tabcolsep}{3mm}{
			\begin{tabular}{ c c c c}
				\toprule
				\multirow{1}*{Methods}&Precision  &Recall&F1-score \\
				
				\midrule
               
				\multirow{1}*{Adaptive-Seg \citep{Pun}} & 23.02 & 13.27 & 13.46\\
				
				\multirow{1}*{DenseField \citep{Cozzolino}} & 22.23 & 23.63 & 22.60\\
				\multirow{1}*{BusterNet \citep{2018Buster}} &  \textbf{51.25} & 28.20 & 35.34\\
				\multirow{1}*{DOA-GAN \citep{2020DOA}}     & 48.42 & 37.84 & 36.92\\
				
                \multirow{1}*{AR-Net \citep{2020AR}}       & 44.79 & 53.82 & 40.31\\
                \multirow{1}*{CNN-T \citep{CNNT}}          & 47.16 & 53.89 & 43.91\\
				
				\midrule
				\multirow{1}*{Baseline}                  & 37.32	&47.98	&40.75 \\
				\multirow{1}*{Prototype}                 & 39.21	&49.33	&42.37 \\
				\multirow{1}*{Prototype (update)}        & 40.37	&50.84	&43.76 \\
				\multirow{1}*{Spatial}                & 41.68	&52.41	&45.15 \\
				\multirow{1}*{Channel}                      & 42.15	&54.09	&46.09 \\
				\multirow{1}*{IMNet}                     & 42.86	&\textbf{55.79}	 & \textbf{46.85} \\
				\bottomrule 
		\end{tabular}}
	\end{table}

Table \ref{tab:uscisi}, Table \ref{tab:casia} and Table \ref{tab:comofod} show the comparison experiment data of our proposed IMNet and Figure \ref{fig:view} shows the visual mask. We can find that IMNet is superior to comparison methods.  Specifically, F1 obtained by IMNet is higher than the sub-optimal framework on USC-ISI CMFD, CASIA v2.0, and the CoMoFoD dataset by $2.34\%$, $1.71\%$, and $2.94\%$.  This is because the object-level modelling we proposed ensures the structure integrity of the detection target. In addition, inconsistent areas between the coarse similar feature and similar prototype are further detected and are set as supplementary to the former refining the detected results effectively.

\subsection{Ablation experiment}
Aiming at verifying the effectiveness of each module of IMNet, we set up some ablation experiments:
\begin{itemize}
    \item Baseline: Detection of similar regions only through convolution and self-correlation calculations.
    \item Prototype: Add prototypes
    \item Prototype (update):  Update the prototype four times per iteration.
    \item Spatial: Mine inconsistencies only by spatial location.
    \item  Channel: Mine inconsistencies only by spatial location as well as corresponding channel elements.
    \item IMNet: Adaptive fuse coarse similar regions and inconsistent representation.
\end{itemize}

Table \ref{tab:uscisi}, \ref{tab:casia}, and \ref{tab:comofod} show the ablation experiment data generated by IMNet.         After introducing the prototype, we carry out experiments on single updates and multiple updates respectively.         The experimental data shows that the optimal effect can be obtained after updating four times in one iteration. In addition, the introducing correlation calculation of the corresponding channels in the Inconsistency Mining module leads to the F1 metric on the USC-ISI dataset increasing by $1.49\%$.

\section{Conclusion}
In this paper, we propose an innovative framework, IMNet, which can detect object-level targets with complete structure by introducing SRP and TRP.         In addition, we propose an Inconsistency Mining module to extract the suspect region between the coarse similar region generated by self-correlation and the updated similarity prototype, which is utilized as supplementary to refine the detection target. Compared with the most advanced approaches, the IMNet we proposed can achieve optimal detection results.

\begin{acks}
This work was supported in part by the National Natural Science Foundation of China (U23A20320),  the Central Government Guided Local Science and Technology Development Fund, China (YDZX2022028), and the China Scholarship Council.
\end{acks}

\bibliographystyle{ACM-Reference-Format}
\bibliography{sample-base}
\appendix

\end{document}